\documentclass{article}

\usepackage{neurips_2021}

\usepackage[utf8]{inputenc} 
\usepackage[T1]{fontenc}    

\usepackage{url}            
\usepackage{booktabs}       
\usepackage{amsfonts}       
\usepackage{nicefrac}       
\usepackage{microtype}      

\usepackage{amsfonts}       
\usepackage{amsmath}
\usepackage{amssymb}
\usepackage[nodisplayskipstretch]{setspace} 
\usepackage[pdftex]{graphicx}    
\newcommand{\mli}[1]{\mathit{#1}}
\usepackage{xcolor}         
\usepackage{hyperref}       
\title{Pastprop-RNN: improved predictions of the future by correcting the past}

\author{%
  André Baptista \\
  Faculdade de Engenharia\\
  Universidade do Porto \\
  \texttt{up201505375@fe.up.pt} \\
  \And
  Yassine Baghoussi \\
  INESC TEC\\ Faculdade de Engenharia\\
  Universidade do Porto \\
  \texttt{baghoussi@fe.up.pt} \\
  \And 
  Carlos Soares \\
  Fraunhofer AICOS and LIACC\\
  Faculdade de Engenharia\\
  Universidade do Porto \\
  \texttt{csoares@fe.up.pt} \\
  \And 
  Jo\~{a}o Mendes-Moreira \\
  INESC TEC \\ Faculdade de Engenharia\\
  Universidade do Porto \\
  \texttt{jmoreira@fe.up.pt} \\
  \And 
  Miguel Arantes \\
  Inovretail \\
  \texttt{miguel.arantes@inovretail.com} \\
}

\begin{document}

\maketitle

\begin{abstract}

Forecasting accuracy is reliant on the quality of available past data. Data disruptions can adversely affect the quality of the generated model (e.g. unexpected events such as out-of-stock products when  forecasting demand). We address this problem by pastcasting: predicting how data should have been in the past to explain the future better.
We propose Pastprop-LSTM, a data-centric backpropagation algorithm that assigns part of the responsibility for errors to the training data and changes it accordingly.
We test three variants of Pastprop-LSTM on forecasting competition datasets,  M4 and M5, plus the Numenta Anomaly Benchmark.
Empirical evaluation indicates that the proposed method can improve forecasting accuracy, especially when the prediction errors of standard LSTM are high. It also demonstrates the potential of the algorithm on datasets containing anomalies.
\end{abstract}

\section{Introduction}
\label{intro}
In Machine Learning (ML), data is at the core of any modelling task. With the increase in the amount of data and its complexity, modelling this data became a challenge. ML models are often evaluated by accuracy, fairness and robustness. The quality of a model is, usually, a reflection of the quality of the underlying data, e.g. noises, anomalies, and additional adversarial perturbations. Therefore, one of the promising ways to improve the quality of the models is to improve the quality of the dataset \cite{renggli2021data}. However, depending on the algorithms used, the description of data quality varies. For instance, some researchers have demonstrated the positive influence of adding noise when classifying the data using deep learning \cite{DBLP:journals/corr/RolnickVBS17}. That is to say, the data quality does not depend only on the data itself but also on the learning nature of the underlying model.  Therefore, giving the models the power to change the data seems to be another promising way to improve the quality of our models.
Forecasting is an ML task that counts several algorithms that are highly dependent on the data quality \cite{SINGH19991389} such as Recurrent Neural Networks (RNNs) \cite{pascanu2013difficulty}. Attempts to improve the quality of these models mainly revolved around changing the architecture of the network (\cite{pascanu2013difficulty};\cite{hochreiter1997lstm}) or improving the optimization component of the algorithm \cite{zaremba2014recurrent}. Deep learning-based RNN such as LSTM performs better in long term prediction problems \cite{siami2019comparative} when compared to traditional statistical methods. In contrary to this latter, RNNs can extract non-linear components from the data \cite{siami2018comparison} which makes them crucial with today's increase of complexity in data. As a side effect, this makes the model sensitive to the occurrence of disruptions in data. For instance, in retail, past sales data are used to estimate future demand. Unexpected increase or decrease of sales of a certain product caused by lack of stock or customer behaviour will disturb complex sequential models such as RNN. During the learning, the recurrent network will try to map this information, which can mislead the overall understanding of data. It is worth noting that an increase and decrease in sales for a particular period (e.g. Christmas season) should not be considered disruptions. The reasoning behind this is that, despite constituting an outlier, the phenomenon is expected and likely helpful at explaining future time series values. Therefore, those values will probably be bad predictors of demand in the future. Such is the case where sales are not representative of demand. For example, when a product goes out of stock for some time, the registered sales for that product will not consider the number of lost sales opportunities. In other words, the observed value of sales is limited by the available stock (i.e.~ an upper bound), and the true demand is outside the measurable range. These disruptions effectively decrease the quality of the input data. 
Finally, we differentiate disruptions from anomalies by the fact that disruptions can be valid and true but misleading. 

Motivated by these issues, we address the general problem of data disruptions by making RNN detect unexpected events in past data and reconstruct them accordingly. We refer to this as \emph{pastcasting}. We propose \emph{Pastprop}, which consists of modifying the backpropagation algorithm of an RNN variant, namely, Long Short Term Memory (LSTM) \cite{hochreiter1997lstm}, to reconstruct the disruptions in data, so reducing their effect on forecasting accuracy. The idea is to make the learning process estimate the contribution of training data to the outputted errors. In other words, the responsibility of errors is shared not only among network weights but also with the data. Hence, instances in the training data should also be updated in the direction that minimizes error.

The main contributions of our work: 
\begin{itemize}
    \item a data-centric RNN using an adaptive backpropagation to improve data quality as part of the learning process; 
    \item three variants of Pastprop;
    \item empirical study of the approach on benchmark datasets and real world dataset.
\end{itemize}

\section{Related Work}
\label{gen_inst}

For some machine learning algorithms, short data size with high quality can lead to better models than data with large size and low quality. In spite of this, the task of cleansing data and learning models are essentially done independently.

\subsection{Data-centric Machine Learning} 
Data-centric Machine Learning focuses on improving the quality of the machine learning models through the improvement of the quality of data (e.g. \cite{renggli2021data}). Often data used in machine learning are interconnected, i.e. every input depends on others (e.g. timeseries). Hence, issues on one input may harm the overall understanding of the model. To eliminate problems in data, practitioners usually precede the learning with data preparation. For example, anomaly detection (\cite{7178320};\cite{inproceedingsADLSTM}) or/and data reconstruction \cite{JEONG2019100991}. The detection process used is, often, data-driven. Data-driven pre-processing uses statistics derived from the data to eliminate or detect a problem. Most of the time, these techniques change the data independently from the model to be used. Although effective, these techniques can fail as many problems in data do not exceed any statistical boundaries – for instance, they may have values that are statistically normal, but are unusual at the specific time that they occur \cite{geiger2020tadgan}. Also, data effects on the model depends on each algorithms (See: \cite{DBLP:journals/corr/RolnickVBS17}). On the other hand, Data-centric ML suggests using feedback from a ML model to detect problems in data \cite{renggli2021data}. For example, RNN-based Anomaly Detection methods. Authors in \cite{malhotra2015long} have used a stacked LSTM network to detect deviations from normal behaviour. They achieve this by combining LSTM with recurrent hierarchical processing layers. \cite{Hayton5} and \cite{ma6}) used the errors on the predictions to detect novelty. They apply a threshold on the prediction error sequences or the transformed error sequences. In the literature, two methods are commonly used to determine the threshold. The first method consists of choosing a fixed threshold using Gaussian distribution of errors \cite{ malhotra2016lstm},\cite{ 7486356}. Because of the fixed nature of the threshold, these methods are often violated. In (\cite{ doi:10.1080/00224065.1986.11979014};\cite{hundman2018detecting}), authors used exponential weighted moving average (EWMA) for a dynamic approach of selecting thresholds. They apply EWMA on the error sequence, obtaining the smoothed error sequence based on which the dynamic threshold is selected. This method does not require prior knowledge about error distribution nor anomalous samples to determine the threshold. In the majority of the existing RNN-based AD methods, the thresholds are set based on, solely, one kind of information which describes the characteristic of the individual anomaly candidate: either the error sequence \cite{ malhotra2016lstm},\cite{ 7486356}, or the transformed error sequence \cite{hundman2018detecting}, \cite{2018JPhCS1061a2012Z}. Although proven to act better than methods using threshold directly on the data, the majority of these methods are limited due to lack of consideration of the data changes over time, so this leads to false alarms. RNN-based methods that address data quality improvement are limited, and most of them were proposed to deal with missing data such as \cite{nature2018}, \cite{choi2015}. 
Although ML models have been already involved in data cleansing, most methods tend to separate the task of correcting data from the learning. In this work, we assume that the models need to learn and correct data simultaneously. Hence, allowing the model to avoid data errors that may propagate during the time.
\subsection{Long Short Term Memory}

Long-Short Time Memory (LSTM) is a type of Recurrent Neural Network (RNN). RNNs differ from feed-forward neural networks due to the recurrent connections which allow them to learn from sequential data. They attempt to model and remember temporal dependencies in sequences. However, RNNs have been criticized due to the vanishing and exploding gradient problem \cite{hochreiter1991untersuchungen}. Because of this, RNNs are not suited to learn long dependencies in time series data. In fact, they are not effective at learning relationships in data more than 5 to 10 time steps apart \cite{kn:Staudemeyer2019}. \cite{hochreiter1997long} proposed LSTM, an RNN based on a gating mechanism. 

The LSTM architecture is composed of blocks, each of them containing an input gate, forget gate, output gate (Eq. \ref{eq:lstm1}- Eq.\ref{eq:lstm3}) and memory cell (Eq. \ref{eq:lstm5}). The purpose of the gates is to control which inputs are more relevant. A gate is sequence of nodes with sigmoid activation functions that are connected to their inputs through learnable weights. The memory cell holds information from previous inputs. The forget gate (Eq.~\ref{eq:lstm3}) learns which information should be kept in memory. Finally, the output gate (Eq.~\ref{eq:lstm2}) controls which information should be passed to the following block. 

The learning of an LSTM is done through a forward and backwards pass. In the forward pass, the input $x_t$ is concatenated with the hidden state $h_{t-1}$ from the last block plus a bias value $b_i$. This is called the hidden input (Eq. \ref{eq:lstm1}). The hidden state from previous time steps $h_{t-1}$ is transported to all the gates, which are then sigmoid activated (Eq. \ref{eq:lstm1}- Eq.\ref{eq:lstm3}). The cell state is calculated by adding two parcels (Eq. \ref{eq:lstm5}). The first is the product of the new candidate memory cell $\overset{-}{C_t}$ and the input gate's activation $i_t$. The second parcel is the product of the previous cell state $C_t$ with the forget gate's activation $f_t$. Finally, the hidden output is obtained by multiplying the \texttt{tanh} of activated cell state $C_t$ with the output gate's activation $o_t$ (Eq. \ref{eq:lstm6}). The new hidden state is then passed to the next block, which will also receive a new input sample. The process is repeated until the last block of the LSTM's architecture is reached. There, the hidden state $h_t$ goes through a final layer of weights to produce the predicted output (Eq. \ref{eq:lstm7}) (or through multiple layers if we want to predict more than one value). After this, the predicted output is compared to the real output and an error measure is obtained. The backwards pass starts by deriving the error with respect to the final layer's weights. This gradient will be used to modify them accordingly. Updating the input, forget and output gates requires using the chain rule of derivatives. First, the derivative of the hidden output with respect to the error is calculated. This is used to calculate the output gate's and the cell state's derivative. The former leads to the input gate's derivative. In conjunction with the previous cell state, it also leads to the forget gate's derivative. This process is repeated for all of the LSTM's blocks. The gradient deltas that will update the gates at each block are finally obtained by multiplying that block's hidden output by each of the gate's derivatives.
\begin{subequations}
\setstretch{0.2} 
    \begin{equation}
    i_t = \sigma(W_i \cdot h_{t-1} + V_i \cdot x_t + b_i)
    \label{eq:lstm1}
    \end{equation}
    \begin{equation}
    o_t = \sigma(W_o \cdot h_{t-1} + V_o \cdot x_t + b_o)
    \label{eq:lstm2}
    \end{equation}
    \begin{equation}
   f_t = \sigma(W_f \cdot h_{t-1} + V_f \cdot x_t + b_f)
    \label{eq:lstm3}
    \end{equation}
    \setstretch{0.2} 
    \begin{equation}
    \overset{-}{C_t} = tanh (W_c \cdot h_{t-1} + V_c \cdot x_t + b_c)
    \label{eq:lstm4}
    \end{equation}
    \begin{equation}
    C_t = i_t \cdot \overset{-}{C_t} + f_t \cdot C_{t-1}
    \label{eq:lstm5}
    \end{equation}
     \begin{equation}
    h_t = o_t \cdot tanh (C_t) 
    \label{eq:lstm6}
    \end{equation}
    \begin{equation}
      z_t = h_t
    \label{eq:lstm7}
    \end{equation}
\end{subequations}
\section{Pastprop}
\label{headings}
As discussed earlier, correction of data issues is usually done separately from learning the model. We propose Pastprop, an adaptation of backpropagation for that purpose \footnote{https://kdd-milets.github.io/milets2020/papers/MiLeTS2020\_paper\_18.pdf}. It consists of reworking the backpropagation algorithm so that error responsibility is extended to data. The premise is that those data issues (e.g. anomalies) have a significant responsibility for the errors made by the model that they are used to train. Furthermore, it is assumed that it is not possible to change the model weights to adequately compensate for those errors. As such, similarly to the network weights, input data is corrected in the direction that minimizes the error of the model it is being used to train, with a magnitude proportional to its contribution to that error. The goal is that the corrections improve the overall quality of the training data and, thus, lead to better models.  

\subsection{Calculating deltas}

The key variables used by pastprop are the deltas. In a similar manner to how the input, forget and output gate's gradients are obtained, we can also calculate a gradient for the LSTM's hidden input at each block. This is done by simply multiplying the gate's derivatives with their respective weights. Since the hidden input $i_t$ is composed of the input sample $x_t$, previous hidden state $h_{t-1}$ and a bias value $b_t$, we subset the gradient to just the part that respects to the input sample. The deltas to be added to the input samples are calculated from this gradient and the \emph{data correction rate} constant, a Pastprop specific hyperparameter which serves as a learning rate for data instead of weights. The following is the expression we are looking for:

\begin{equation}\label{eq:errorhint}
\begin{gathered}
\frac{\partial \mli{Error}_{t}}{\partial \mli{hin}_{t}} = \frac{\partial \mli{Error}_{t}}{\partial (\mli{hin}_{t} W_{i})} W_{i} + \frac{\partial \mli{Error}_{t}}{\partial (\mli{hin}_{t} W_{f})} W_{f} 
+ \frac{\partial \mli{Error}_{t}}{\partial (\mli{h}_{t} W_{o})} W_{o} \\
+ \frac{\partial \mli{Error}_{t}}{\partial (\mli{hin}_{t} W_{g})} W_{g}
\end{gathered}
\end{equation}

Since the hidden input is composed of the input sample, previous hidden output and a bias value

\begin{equation}\label{eq:hin}
    \mli{hin}_{t} = \{X_{t}, \mli{hout}_{t-1}, b\}
\end{equation}

We subset the gradient to just the part that respects to the input sample.

\[ \frac{\partial \mli{Error}_{t}}{\partial \mli{hin}_{t}} [X_{t}] \]

\subsection{Pastprop variants}
\label{pastprop_variations}
The following subsections explain the implementation of three different Pastprop variants.
\subsubsection{Epoch-wise Pastprop}

In this version, the corrections are applied to the whole time series -- all at once -- after each epoch has been completed. During an epoch, the "deltas" to be added to each data sample are calculated and stored until the end of the epoch. Right before starting the new epoch, the deltas and time series are added together. This means that every epoch deals with a different version of the data. It also means that the first epoch produces exactly the same weights that a normal LSTM would (in case the initial weights are the same).

\begin{figure}
  \centering
       \includegraphics[width=0.65\linewidth]{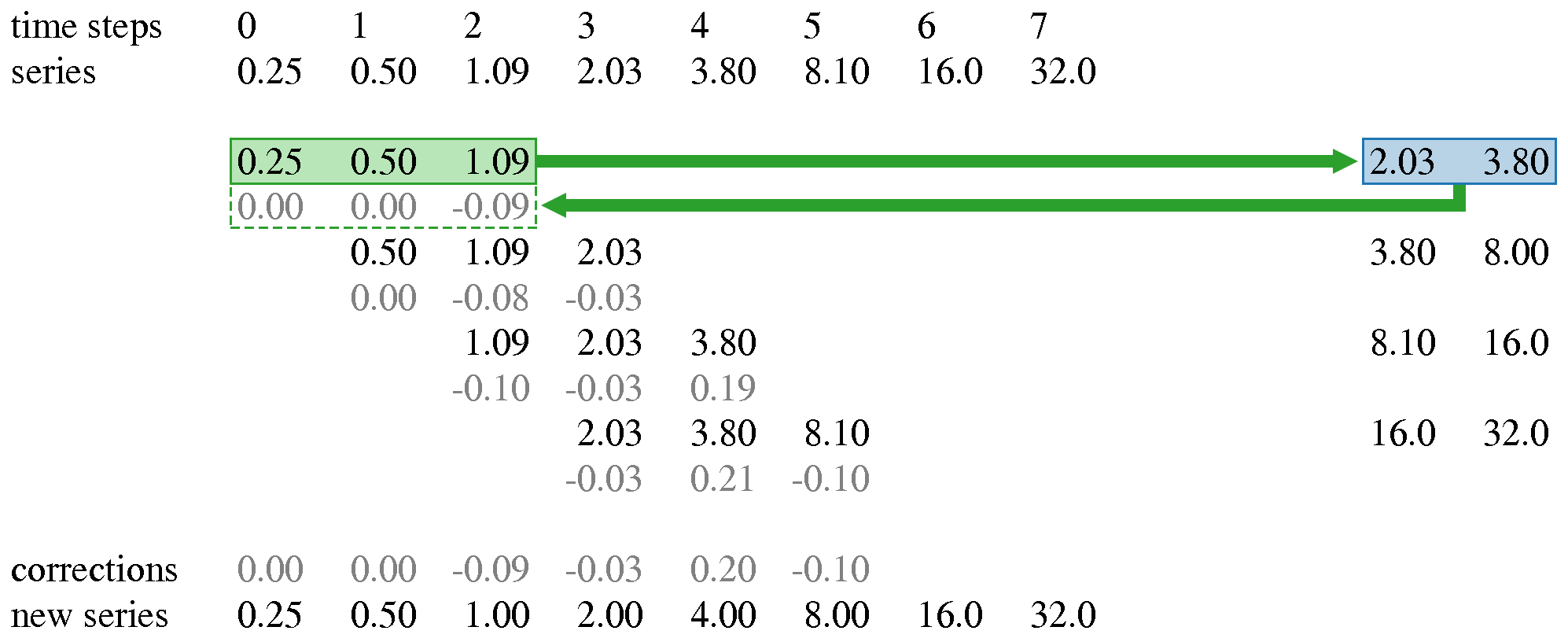}
  \caption{Simplified illustration of how Epoch-wise Pastprop works}
  \label{fig10}
\end{figure}

Figure \ref{fig10} illustrates the described mechanism in a simplified way. We can see that most time steps have overlapping deltas associated with them. In those cases, an average of the corresponding deltas (not their sum) is used to build the final corrections array. This is done to make the amount of data corrections less dependent on the sample size.

\subsubsection{Instance-wise Pastprop}

This variant differs from the "Epoch-wise" one in the sense that deltas are added to data over the course of epochs. As soon as they are obtained, deltas are first adjusted to compensate for overlapping and then added to samples. 

Contrary to the Epoch-wise variant, each epoch deals with progressively different time series. For this reason, the weights differ from those produced by an LSTM right from the first epoch.

\subsubsection{Selective Pastprop}

This variant is built upon the Epoch-wise variant and introduces a new feature. Pastprop should only try to correct data issues that affect the quality of the model. We introduce the \emph{correction threshold}, $t_c$, that prevents the algorithm from making small changes, which are probably useless. For the same reason, we introduce a neighborhood mechanism, that increases the likelihood of a data point being corrected if its neighbors also are estimated to be affected by the same data issue. This mechanism is controlled by the \emph{neighborhood size}, $s$, parameter. Since standard NN use random initial weights, data correction should only start after some model learning (i.e. weight adaptation) has taken place. We introduce \emph{epoch embargo}, $e$, an additional hyperparameter, that determines the number of epochs that are executed before the data correction mechanism starts. 
Keeping the time series unchanged for the first few epochs is an attempt at reducing the impact of the LSTM's random weight initialization on data corrections. As mentioned previously, the calculation of deltas is dependent on weights. This implies that deltas obtained early in the first epoch are based on very close to random weights. Waiting for the completion of some initial epochs $e$ before applying corrections makes it more likely for the weights to reach a meaningful state. The rationale behind the thresholding is to avoid unnecessary changes to data. Maintaining only the most significant deltas from the corrections array is an effort to achieve this. Moreover, the inclusion of neighbor deltas in the ranking system incentivizes the preservation of deltas associated with anomalous zones. In turn, single point anomalies are not as targeted by the system.

\section{Experimental setup}
\label{others}

In order to evaluate Pastprop, we have compared its performance with standard LSTM on multiple timeseries. In these experiments, it is assumed that the algorithms receive an univariate time series as training data. During the learning process, multi time step samples are used to predict multi time step labels. At the end of training, the outputs are both the network's weights and the corrected time series. Since, Pastprop has been designed for improving data quality, we have, also, tested the ability of the algorithm on reconstructing anomalies that were systematically introduced in M5 dataset.

\subsection{Research questions}

We identify the following as the main research questions of our empirical study:

\begin{itemize}

\item Can Pastprop improve time series forecasting accuracy ?
\item How effective is Pastprop at reconstructing anomalies in data ?
\item Do other algorithms benefit from using the improved training data ?
\end{itemize}{}

\subsection{Hyperparameters}

The LSTM related hyperparameters required for the Pastprop methods are the: sample size = 5; label size=1; number of hidden units for each LSTM gate = 200; learning rate=0.001; number of epochs=50.
The only Pastprop specific hyperparameter common to all implementations is the data correction rate $t_c$. The Selective version also needs the: number of waiting epochs $e$; number of highest ranked deltas to keep on the corrections array; number of previous and subsequent time steps to consider when ranking deltas.

\subsection{Methodology}

The pastprop variants were implemented based on a publicly available Python implementation of an LSTM \cite{batched}. For a given time series and a complete set of parameters, a single experiment consisted of performing multiple operations. The first step was to generate random initial weights compatible with the problem at hand. For the sake of fairness in comparison, all tested methods worked with these same starting weights. The methods were a simple LSTM, all three Pastprop variants tuned with equal hyperparameters. Besides forecasting accuracies on test data, a few other results were registered. These included similarity measures between each corrected series and the original data, the correlation between the timeseries length and accuracy, and finally the correlation between the accuracy of Pastprop variants and the scale of errors made by standard LSTM.
\begin{table}
\centering
\caption{Descriptive analytics of M4 and M5 Forecasting Competition Datasets.}
\label{tab:1}
\small
\begin{tabular}{lccccccc}
                   \cline{2-8}
                   & M5                        & \multicolumn{6}{c}{M4} \\                       
\cline{2-8}
                   & \multicolumn{1}{l}{Hobbies} & Day  & Hour  & Month & Quarter & Year & Week \\ \cline{1-8} 
Timeseries         & 205                       & 100  & 100   & 100   & 100     & 100  & 100  \\ 
Average Length     & 1939                      & 558  & 700   & 366   & 90      & 42   & 1255 \\ 
Mean               & 0.9147                    & 2978 & 19935 & 4222  & 4154    & 4230 & 3925 \\ 
Standard Deviation & 1.419                     & 688  & 3621  & 1160  & 838     & 1456 & 1998\\
\hline
\end{tabular}
\end{table}

\begin{table}
\centering
\small
\caption{Descriptive analytics of Numenta Anomaly Benchmark Datasets.}
\label{tab:2}
\begin{tabular}{lccccc}
\cline{2-6} 
                     & \multicolumn{5}{c}{NAB}                    \\ \cline{2-6} 
\multicolumn{1}{c}{} & RKC   & Traffic & AWS     & Art   & Tweets \\ \cline{1-6} 
Timeseries           & 7     & 7       & 17      & 9     & 10     \\ 
Average Length       & 9872  & 2236    & 3980    & 4030  & 15854  \\ 
Mean                 & 40.18 & 124.73  & 742771  & 27    & 42.74  \\ 
Standard Deviation   & 6.88  & 125.03  & 3.1e+06 & 26.84 & 98.74 \\ \hline
\end{tabular}
\end{table}


\subsection{Baseline algorithms}

The procedure just described was the main methodology developed to conduct our study. However, additional experiments were needed to study all research questions. Other algorithms, traditionally used in forecasting tasks, were chosen as references. Those were the following:

\begin{itemize}
\item Autoregressive integrated moving average (ARIMA)
\item Exponential smoothing state space model (ETS)
\end{itemize}{}

The baseline algorithms were executed with anomalous data. Their results were compared to the forecasting accuracies obtained by the Pastprop implementations. Lastly, the same algorithms were run with the corrected time series outputted by Pastprop. The results were then compared to those obtained earlier, while using the original anomalous data.

\subsection{Datasets}
\label{Datasets}
The research was performed on 855 timeseries (Tables~\ref{tab:1} and~\ref{tab:2})  taken from M4 and M5 forecasting competition datasets and Numenta Anomaly Benchmark \cite{lavin2015evaluating}. Parameters were kept constant throughout all experiments. For each train data from M4 datasets, its corresponding test data was used. As for M5 and NAB, the training data consisted of the first 70\% of the time series and the last 30\% were used for testing. Furthermore, the learning rate was always 0.001 and the number of hidden units was always 200. Lastly, all time series were normalized between 0 and 1.

\paragraph{Artificial Anomalies}
In order to study the ability of Pastprop on reconstructing missing data/anomalies, we have artificially inserted anomalies to M5 datasets. Anomalies were inserted with three different levels of magnitude. The first, level 0, simply substitutes the data by a sequence of zeros. The other two levels function in a different way. The anomaly zone is divided in evenly sized chunks. It is randomly chosen whether each chunk should add or subtract from the original data. The value of change at a given time step is the maximum between 0.1 and a percentage of the time series' original value. The possible percentages are 25 and 50 which correspond to the magnitude levels 25 and 50. The minimum absolute change value of 0.1 was chosen in the context of data being normalized between 0 and 1.

\subsection{Evaluation Metric}
We measure the performance of standard LSTM and Pastprop using the commonly used metric MSE.
\begin{equation}
    MSE = \frac{1}{n}\sum_{t=1}^{n}e_t^2
\end{equation}
As the timeseries in NAB datasets have different scales, we have normalized the MSEs for comparison by dividing the values by the average of target values.

\section{Results}
Pastprop evaluation was, mainly, performed to test whether giving part of responsibility of errors to data can  improve forecasting on time series with and without known problems in data. Additionally, Pastprop has been tested on its reconstruction ability. Finally, we analyzed the performance of ARIMA and ETS algorithms using the reconstructed data and compare them to the performance of Pastprop.

\begin{table}
\centering
\caption{Mean Squared Error of standard LSTM and Pastprop variants on M4 and M5 competition dataset.}
\label{tab3}
\small
\begin{tabular}{lccccccccc}
            \cline{2-8}
            & M5             & \multicolumn{6}{c}{M4}                                                                              & Mean           & \begin{tabular}[c]{@{}l@{}}Standard\\ Deviation\end{tabular} \\ \cline{2-8}
            & Hobbies          & Day         & Hour         & Month        & Quarter      & Year         & Week         &                &                                                              \\ \cline{1-10} 
LSTM        & 0,028          & 0,040          & \textbf{0,032} & 0,023          & 0,125          & 0,294          & 0,019          & 0,1014         & 0,08                                                         \\ Epoch-wise     & 0,028          & 0,073          & 0,049          & 0,032          & 0,163          & \textbf{0,046} & 0,326          & 0,109          & 0,103                                                        \\ 
Instance-wise & 0,028          & 0,057          & 0,079          & 0,041          & 0,206          & 0,335          & 0,295          & 0,128          & 0,148                                                        \\ 
Selective   & \textbf{0,024} & \textbf{0,017} & 0,137          & \textbf{0,017} & \textbf{0,029} & 0,100          & \textbf{0,014} & \textbf{0,049} & \textbf{0,048}                                               \\ \hline
\end{tabular}
\end{table}

\begin{table}
\centering
\caption{Mean and standard deviations of Normalized Mean Squared Error of standard LSTM and Pastprop variants on Numenta Anomaly Benchmark dataset.}
\label{tab4}
\small
\begin{tabular}{lccccc}
            \cline{2-6}
              & \multicolumn{5}{c}{NAB}                                                                                                                           \\ \hline
              & Art                        & AWS                         & RKC                        & Traffic                        & Tweets                      \\ \hline
LSTM & \textbf{0,0049 $\pm$ 0.02} & 0,0024 $\pm$ 0.013 & \textbf{0,016 $\pm$ 0.005} & 0,0104 $\pm$ 0.013          & 0,0005 $\pm$ 0.001          \\ 
Epoch-wise      & 0,012  $\pm$ 0.063         & 0,020 $\pm$ 0.02            & 0,034 $\pm$ 0.048          & 0,014 $\pm$ 0.015           & \textbf{0,0002 $\pm$ 0.000} \\ 
Instance-wise   & 0,007 $\pm$ 0.982          & \textbf{0,0016$\pm$ 0.013}  & 0,410 $\pm$ 0.003          & \textbf{0,0102 $\pm$ 0.013} & 0,0005 $\pm$ 0.001          \\ 
Selective     & 0,017 $\pm$ 0.159          & 0,007$\pm$ 0.017            & 0,079 $\pm$ 0.021          & 0,014 $\pm$ 0.015           & 0,0008 $\pm$ 0.002          \\ \hline
\end{tabular}
\end{table}

\begin{figure}
  \centering
      \includegraphics[width=1\linewidth]{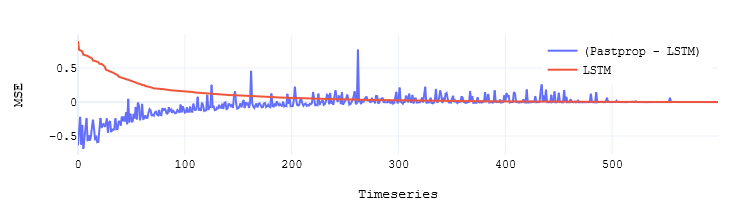}
  \caption{The evolution of Pastprop errors gains with corresponding Mean Squared Errors of LSTM. The red line shows the standard LSTM errors; The blue line shows the decrease produced by Pastprop over standard LSTM.}
  \label{fig:4}
\end{figure}

\subsection{Can Pastprop improve time series forecasting accuracy?}
For a given data correction rate of 1, Selective Pastprop has been the best performing among the variants with 48\% MSE decrease over standard LSTM (Table~\ref{tab3}). The performance of other variants was either equal to standard LSTM or worst. This can be explained by the fact that Epoch-wise and Instance-wise are both correcting data without being supervised. That is to say, these variants are taking every disruptions, be them small or large, into account which may lead to error accumulation and propagation during time. Selective Pastprop minimizes this side effect through the thresholding strategy explained previously. Also, we have used a correction rate of 1, a smaller rate would provide a better performance in these variants. Fianlly, a more sophisticated backend implementation of LSTM would provide more confidence in assigning responsibility of errors to the input sequence.

As for NAB, Instance-wise Pastprop performed better by winning over LSTM and other variants on two datasets AWS and Traffic. Epoch-wise Pastprop slightly decreased the errors in Tweets dataset. In contrary to the first experiment, Selective Pastprop was the worst performing Pastprop (see: Table \ref{tab3}).

Motivated by \cite{cerqueira2019machine}, the correlation between the timeseries lengths and accuracy of Pastprop was measured. It is commonly known that LSTM performance is highly dependent on the input sequence length. In our experiments, Pearson standard correlation coefficient between LSTM predictions accuracy and the timeseries length shows 41\% positive. This means that the larger the series are, the better is the accuracy. In the other hand, we had 33,79\%, 10,5\% and 11,51\% positive for Epoch-wise, Instance-wise and Selective Pastprop resp. Further investigations have been carried out using the best performing Pastprop (i.e. Selective). The evolution of the gains over M4 dataset shows that the gains are higher when LSTM MSEs are higher (see: Figure \ref{fig:4}).

\subsubsection{Sensitivity Analysis}
\label{sens_analysis}
The previous experiments were run using a correction rate equals to 1. In order to test the sensivity of Pastprop variants to this hyperparameter, we run another experiments using M5 dataset with a correction rate of 0.01 (see: Table \ref{tab:sample-table}). In addition, we additionally compared the performance of these variants against ARIMA, ETS, and  standard LSTM using MSE. ARIMA was the best performing algorithm with an average MSE of 0.01435 followed closely by Instance-wise Pastprop achieving 0.01449, as well as Epoch-wise Pastprop with 0.01474. Both of these variants were able to improve the accuracy achieved by LSTM in 9 out of 12 time series, with a combined average MSE decrease of 13\%. In contrary to results from the first experiment, where Selective Pastprop gave the best score, in this experiment Selective provided a lower but a considerable accuracy gain of 3\%. It still overpassed LSTM results in 8 occasions. Moreover, Pastprop is highly dependent on the model performance. Since the corrections are only made after the model predicts, a very bad performance will lead to insignificant corrections. Thus, it is recommended that Pastprop uses a good implementation of LSTM on its backend.

\subsection{Do other algorithms benefit from using the corrected time series as training data?}

The process of studying this question was based on a selected subset of time series from M5 dataset with artificially inserted anomalies. The ARIMA and ETS algorithms were used for comparison. Each of these algorithms produced one forecast for each of those anomalous data. Then, we gathered the respective Pastprop corrected M5 timeseries that were produced during the main experiments. These timeseries were given to ARIMA and ETS and their forecasts were registered. We also retrieved the accuracies of the LSTM with the corrected series as training data that had been registered on the main experiments. Finally, the algorithms' results per time series were averaged by Pastprop version.

The results showed that the corrections had, practically, no effect on the forecasting ability of ETS. Regarding ARIMA, the changes to accuracy were negative yet minimal. The Epoch-wise and Selective Pastprop corrected time series had a negative impact of 0.5\% on ARIMA's forecasting accuracy. The series corrected by Instance-wise Pastprop produced 0.2\% worse forecasts. Finally, the results on LSTM were a bit more significant. This time, on average, all series corrected by Pastprop versions proved to be beneficial over the original anomalous data. The Instance-wise and Selective variants achieved the best results, both with 6,6\% gains, while Epoch-wise Pastprop still achieved 2,5\%. It should be noted that the LSTMs which were given the corrected time series were initialized with the same weights as the Pastprop implementations that produced them. This could be relevant for the achieved results. To conclude, the correction made by Pastprop seem to be necessary for LSTM. Based on these results, we can say that the corrections made by Pastprop are carried out on input that are disturbing the model itself and can not, necessarily, be considered as anomalies.

\subsection{How effective is Pastprop at reconstructing anomalies in data?}

\begin{figure}
  \centering
      \includegraphics[width=0.4\linewidth]{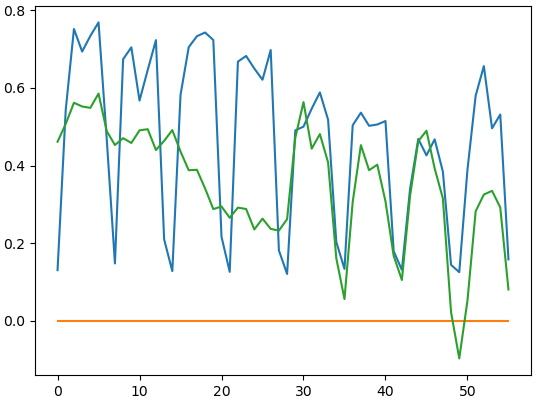}
  \caption{Missing data reconstruction example. Orange line represents the missing data, blue consists of the true values and green are the reconstructions.}
  \label{fig:8}
\end{figure}

The anomaly reconstruction ability was calculated by resorting to two metrics: the distance between the corrections and the original data; the distance between the anomaly and the original data. The former was divided by the latter and the result was subtracted from 1. As such, a positive reconstruction ability indicates the corrections were able to push the anomalous zone closer to the original data. On the other hand, a negative value means the anomaly was accentuated even further. The distance between corrections and original data was also measured while excluding the anomalous part. This will be referred to as the \emph{outside loss}. Therefore we want this value to be as low as possible. When considering other types of data, this measure loses some of its meaning since corrections may be beneficial even outside of the anomaly. Analyzing the experiments on the M5 benchmark datset, the global averages were of 4\% reconstruction ability in the first two Pastprop versions and 1.5\% in the Selective one. The respective outside losses were around 0.01 and 0.0005. Figure~\ref{fig:8} shows one of the runs of a particular experience done on that time series, in which missing data reconstruction achieved 81\%.
\begin{table}
\centering
\small
\caption{Forecasting accuracies (MSE) of the baseline, LSTM and Pastprop algorithms on M5 data that was previously affected by anomalies. Bold and underline are best and second best resp.}

\begin{tabular}{llll|llll}

\hline
M5        & Magn. & ARIMA            & ETS     & LSTM    & Epoch-wise         & Instance-wise      & Selective \\ \hline
Foods     & 0     & 0.01105          & 0.01304 & 0.00892 & \underline{0.00809}          & \textbf{0.00763} & 0.00833   \\ 
Foods     & 50    & 0.01292          & 0.01304 & \underline{0.00823} & \textbf{0.00823} & 0.00847          & 0.00848   \\ 
Hobbies   & 0     & \textbf{0.01059} & 0.01214 & 0.0269  & 0.02083          & \underline{0.0208}           & 0.02582   \\ 
Hobbies   & 50    & \textbf{0.01093} & 0.01214 & 0.02916 & 0.0222           & \underline{0.0215}           & 0.02716   \\ 
Household & 0     & \textbf{0.01252} & 0.01589 & 0.01539 & \underline{0.01351}          & 0.01414          & 0.01356   \\ 
Household & 50    & \textbf{0.01284} & 0.01633 & 0.0137  & 0.01327          & \underline{0.01283}          & 0.01454  \\ \hline
\end{tabular}
\label{tab:sample-table}
\end{table}

\section{Conclusion}
We show that the data quality depends not only on the data itself but also on the underlying machine learning model. In this work, we show that by giving LSTM models the power of changing the data during their learning, they can increase their prediction accuracy. Also, we demonstrate that these changes cannot be generalized for improving other forecasting algorithms. As the algorithm developed in this work was designed to correct disrupting data, the experiments also show the potential of this algorithm in reconstructing anomalies or missing data.
\bibliographystyle{plain}
\bibliography{main.bib}

\begin{thebibliography}{10}

\bibitem{cerqueira2019machine}
Vitor Cerqueira, Luis Torgo, and Carlos Soares.
\newblock Machine learning vs statistical methods for time series forecasting:
  Size matters, 2019.

\bibitem{nature2018}
Zhengping Che, Sanjay Purushotham, Kyunghyun Cho, David Sontag, and Yan Liu.
\newblock Recurrent neural networks for multivariate time series with missing
  values.
\newblock {\em Scientific Reports}, 8, 04 2018.

\bibitem{choi2015}
Edward Choi, Taha Bahadori, and J.~Sun.
\newblock Doctor ai: Predicting clinical events via recurrent neural networks.
\newblock {\em CoRR}, 56, 11 2015.

\bibitem{geiger2020tadgan}
Alexander Geiger, Dongyu Liu, Sarah Alnegheimish, Alfredo Cuesta-Infante, and
  Kalyan Veeramachaneni.
\newblock Tadgan: Time series anomaly detection using generative adversarial
  networks.
\newblock {\em arXiv preprint arXiv:2009.07769}, 2020.

\bibitem{Hayton5}
Paul Hayton, Stan Utete, D.~King, S.~King, P.~Anuzis, and L.~Tarassenko.
\newblock Static and dynamic novelty detection methods for jet engine health
  monitoring.
\newblock {\em Philosophical transactions. Series A, Mathematical, physical,
  and engineering sciences}, 365:493--514, 03 2007.

\bibitem{hochreiter1991untersuchungen}
Sepp Hochreiter.
\newblock Untersuchungen zu dynamischen neuronalen netzen.
\newblock {\em Diploma, Technische Universit{\"a}t M{\"u}nchen}, 91(1), 1991.

\bibitem{hochreiter1997long}
Sepp Hochreiter and J{\"u}rgen Schmidhuber.
\newblock Long short-term memory.
\newblock {\em Neural computation}, 9(8):1735--1780, 1997.

\bibitem{hochreiter1997lstm}
Sepp Hochreiter and J{\"u}rgen Schmidhuber.
\newblock Lstm can solve hard long time lag problems.
\newblock In {\em Advances in neural information processing systems}, pages
  473--479, 1997.

\bibitem{hundman2018detecting}
Kyle Hundman, Valentino Constantinou, Christopher Laporte, Ian Colwell, and Tom
  Soderstrom.
\newblock Detecting spacecraft anomalies using lstms and nonparametric dynamic
  thresholding.
\newblock In {\em Proceedings of the 24th ACM SIGKDD international conference
  on knowledge discovery \& data mining}, pages 387--395, 2018.

\bibitem{doi:10.1080/00224065.1986.11979014}
J.~Stuart Hunter.
\newblock The exponentially weighted moving average.
\newblock {\em Journal of Quality Technology}, 18(4):203--210, 1986.

\bibitem{JEONG2019100991}
Seongwoon Jeong, Max Ferguson, Rui Hou, Jerome~P. Lynch, Hoon Sohn, and
  Kincho~H. Law.
\newblock Sensor data reconstruction using bidirectional recurrent neural
  network with application to bridge monitoring.
\newblock {\em Advanced Engineering Informatics}, 42:100991, 2019.

\bibitem{lavin2015evaluating}
Alexander Lavin and Subutai Ahmad.
\newblock Evaluating real-time anomaly detection algorithms--the numenta
  anomaly benchmark.
\newblock In {\em 2015 IEEE 14th International Conference on Machine Learning
  and Applications (ICMLA)}, pages 38--44. IEEE, 2015.

\bibitem{ma6}
Junshui Ma and Simon Perkins.
\newblock Online novelty detection on temporal sequences.
\newblock In {\em Online novelty detection on temporal sequences}, pages
  613--618, 08 2003.

\bibitem{malhotra2016lstm}
Pankaj Malhotra, Anusha Ramakrishnan, Gaurangi Anand, Lovekesh Vig, Puneet
  Agarwal, and Gautam Shroff.
\newblock Lstm-based encoder-decoder for multi-sensor anomaly detection.
\newblock {\em arXiv preprint arXiv:1607.00148}, 2016.

\bibitem{inproceedingsADLSTM}
Pankaj Malhotra, Lovekesh Vig, Gautam Shroff, and Puneet Agarwal.
\newblock Long short term memory networks for anomaly detection in time series.
\newblock In {\em Long Short Term Memory Networks for Anomaly Detection in Time
  Series}, 04 2015.

\bibitem{malhotra2015long}
Pankaj Malhotra, Lovekesh Vig, Gautam Shroff, and Puneet Agarwal.
\newblock Long short term memory networks for anomaly detection in time series.
\newblock In {\em Proceedings}, volume~89, pages 89--94. Presses universitaires
  de Louvain, 2015.

\bibitem{7178320}
Erik Marchi, Fabio Vesperini, Florian Eyben, Stefano Squartini, and Björn
  Schuller.
\newblock A novel approach for automatic acoustic novelty detection using a
  denoising autoencoder with bidirectional lstm neural networks.
\newblock In {\em 2015 IEEE International Conference on Acoustics, Speech and
  Signal Processing (ICASSP)}, pages 1996--2000, 2015.

\bibitem{7486356}
A.~{Nanduri} and L.~{Sherry}.
\newblock Anomaly detection in aircraft data using recurrent neural networks
  (rnn).
\newblock In {\em 2016 Integrated Communications Navigation and Surveillance
  (ICNS)}, pages 5C2--1--5C2--8, 2016.

\bibitem{pascanu2013difficulty}
Razvan Pascanu, Tomas Mikolov, and Yoshua Bengio.
\newblock On the difficulty of training recurrent neural networks.
\newblock In {\em International conference on machine learning}, pages
  1310--1318. PMLR, 2013.

\bibitem{batched}
{Batched LSTM forward and backward pass}.
\newblock \url{https://gist.github.com/karpathy/587454dc0146a6ae21fc}.

\bibitem{renggli2021data}
Cedric Renggli, Luka Rimanic, Nezihe~Merve G{\"u}rel, Bojan Karla{\v{s}},
  Wentao Wu, and Ce~Zhang.
\newblock A data quality-driven view of mlops.
\newblock {\em arXiv preprint arXiv:2102.07750}, 2021.

\bibitem{DBLP:journals/corr/RolnickVBS17}
David Rolnick, Andreas Veit, Serge~J. Belongie, and Nir Shavit.
\newblock Deep learning is robust to massive label noise.
\newblock {\em CoRR}, abs/1705.10694, 2017.

\bibitem{siami2018comparison}
Sima Siami-Namini, Neda Tavakoli, and Akbar~Siami Namin.
\newblock A comparison of arima and lstm in forecasting time series.
\newblock In {\em 2018 17th IEEE International Conference on Machine Learning
  and Applications (ICMLA)}, pages 1394--1401. IEEE, 2018.

\bibitem{siami2019comparative}
Sima Siami-Namini, Neda Tavakoli, and Akbar~Siami Namin.
\newblock A comparative analysis of forecasting financial time series using
  arima, lstm, and bilstm.
\newblock {\em arXiv preprint arXiv:1911.09512}, 2019.

\bibitem{SINGH19991389}
Sameer Singh.
\newblock Noise impact on time-series forecasting using an intelligent pattern
  matching technique.
\newblock {\em Pattern Recognition}, 32(8):1389--1398, 1999.

\bibitem{kn:Staudemeyer2019}
Ralf~C. Staudemeyer and Eric~Rothstein Morris.
\newblock {Understanding LSTM -- a tutorial into Long Short-Term Memory
  Recurrent Neural Networks}.
\newblock {\em arXiv}, sep 2019.

\bibitem{zaremba2014recurrent}
Wojciech Zaremba, Ilya Sutskever, and Oriol Vinyals.
\newblock Recurrent neural network regularization.
\newblock {\em arXiv preprint arXiv:1409.2329}, 2014.

\bibitem{2018JPhCS1061a2012Z}
Runtian {Zhang} and Qian {Zou}.
\newblock {Time Series Prediction and Anomaly Detection of Light Curve Using
  LSTM Neural Network}.
\newblock In {\em Journal of Physics Conference Series}, volume 1061 of {\em
  Journal of Physics Conference Series}, page 012012, July 2018.

\end{thebibliography}

\end{document}